\documentclass[
]{ceurart}

\usepackage{listings}
\usepackage{float}
\usepackage{bm}
\usepackage[outdir=./]{epstopdf}

\begin{document}

\copyrightyear{2022}
\copyrightclause{Copyright for this paper by its authors.
  Use permitted under Creative Commons License Attribution 4.0
  International (CC BY 4.0).}

\conference{LAK'23: Workshop on Partnerships for Cocreating Educational Content, March 13, 2023, Arlington, TX, USA}

\title{Automated Essay Scoring in Argumentative Writing: DeBERTeachingAssistant}


\author[1]{Yann Hicke}[%
email=ylh8@cornell.edu,
]
\cormark[1]
\fnmark[1]
\address[1]{Cornell University,
  Department of Computer Science}

\author[2]{Tonghua Tian}
\fnmark[1]
\address[2]{Cornell University,
  Department of Operations Research and Information Engineering}

\author[3]{Karan Jha}
\fnmark[1]
\address[3]{Cornell University,
  Department of Mechanical Engineering}
  
\author[3]{Choong Hee Kim}
\fnmark[1]

\cortext[1]{Corresponding author.}
\fntext[1]{These authors contributed equally.}

\begin{abstract}
  Automated Essay scoring has been explored as a research and industry problem for over 50 years. It has drawn a lot of attention from the NLP community because of its clear educational value as a research area that can engender the creation of valuable time-saving tools for educators around the world. Yet, these tools are generally focused on detecting good grammar, spelling mistakes, and organization quality but tend to fail at incorporating persuasiveness features in their final assessment. The responsibility to give actionable feedback to the student to improve the strength of their arguments is left solely on the teacher's shoulders. In this work, we present a transformer-based architecture capable of achieving above-human accuracy in annotating argumentative writing discourse elements for their persuasiveness quality and we expand on planned future work investigating the explainability of our model so that actionable feedback can be offered to the student and thus potentially enable a partnership between the teacher's advice and the machine's advice.
\end{abstract}


\maketitle
\section{Introduction}
ETS e-rater \cite{attali2004automated} is one of many commercial tools available today that can automatically grade essays and hence save a substantial amount of human time. It follows a long lineage of tools that have been created over the past 50 years all traced back to Page's pioneering work on the Project Essay Grader \cite{page1966imminence}. All high school students taking the SAT to get into college or undergraduate students applying to graduate schools with their GRE or GMAT scores will have their essays graded by an Automated Essay Scoring (AES) system. The vast majority of AES software are holistic graders in the sense that they summarize the entire quality of an essay in one single score. The main reason that explains such a trend is the nature of the vast majority of annotated corpora available which has a holistic score associated with it.
\par
In August 2022, Crossley et al. released a dataset somewhat unique of its kind: a large-scale corpus of writing with annotated discourse elements (PERSUADE) indicating their level of persuasiveness \cite{crossley2022persuasive}. The originality of this dataset is what is motivating our work: can we achieve human-level accuracy on the persuasiveness prediction task? And then using this performance can we provide feedback to the student-writer?\\

\section{Related work}
We will outline here work that has been done in Automated Essay Scoring that does not solely focus on holistic scoring.
\subsection{Identifying Argumentative Discourse Structures in Persuasive Essays:} In 2014, Stab and Gurevych \cite{stab2014identifying} developed a corpus of essays and tried to identify the structure of arguments in persuasive essays as well as novel feature sets for identifying argument components and argumentative relations, which was one of the first approaches in the field of argument mining.

\subsection{SVM Regressor for Modeling Argument Strength:} In 2015, Persing and Ng \cite{persing2015modeling} proposed an SVM regressor model to score an essay based on the strength of an argument. This paper also released a human-annotated dataset of 1000 essays publically to stimulate further research. In this dataset, the essays were scored from 1 through 4, higher score indicating a strong argument.

\subsection{Neural Models for Predicting Argument Persuasiveness:} In 2018 Carlile et. al. \cite{carlile2018give} released an argument mining dataset, annotating the arguments within the essay as MajorClaims, Claims, Premises, Support and Attack, as well as scored these sections on the basis of attributes like Specificity, Eloquence, Strength, etc. The same group of people in another paper in 2018 \cite{ke2018learning} proposed a bidirectional LSTM model with attention for providing scores on these metrics (Specificity, Eloquence, Strength, etc.) using a neural network, using the same dataset.
\par
In 2019, Toledo et. al. also released a new dataset annotating arguments on the basis of quality and comparing pair-wise arguments for the stronger argument. They used a BERT-base-uncased model architecture to create word embeddings, with an Argument Classification and an Argument Ranking head. \cite{toledo2019automatic}

\begin{figure}[ht]
\centering
\includegraphics[width=\linewidth]{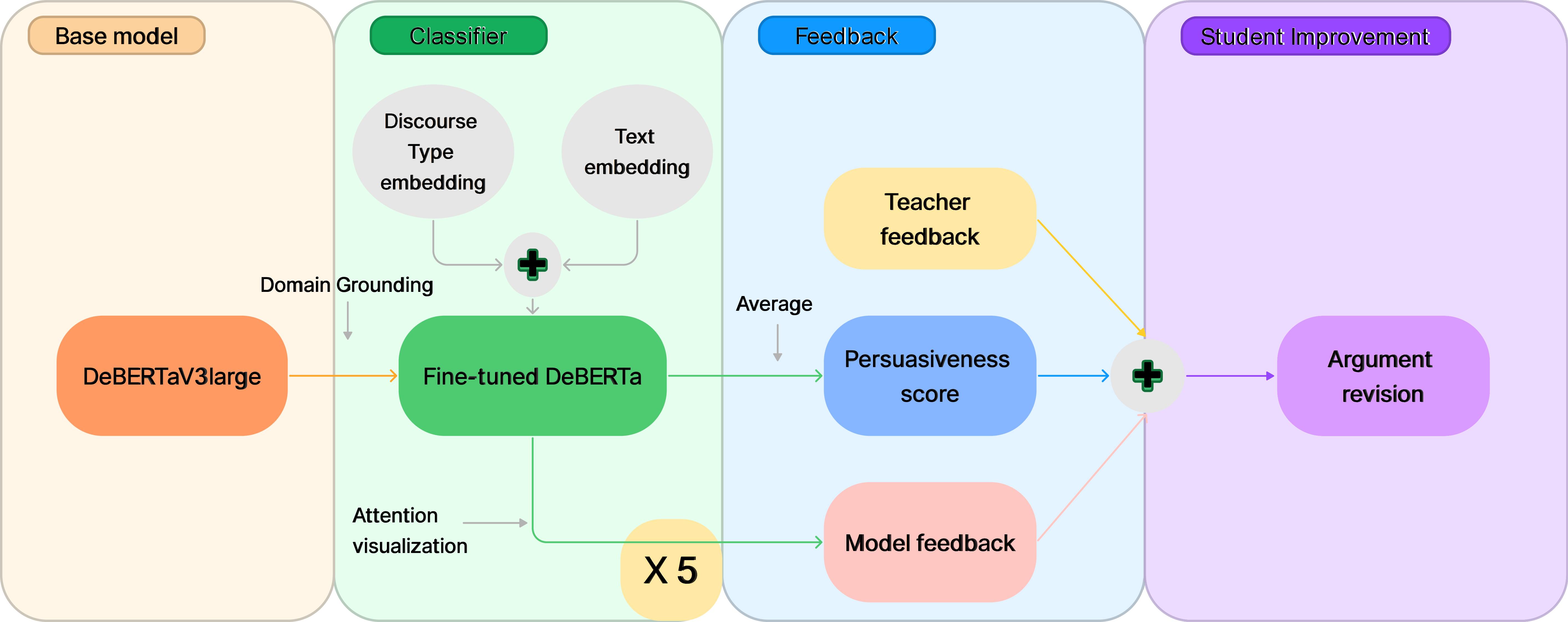}
\caption{The figure presents the architecture of the DeBERTeachingAssistant. The Vanilla DeBERTa-large model is grounded within the domain of the essays using Masked Language Modeling. Five separate classifiers were trained by splitting the training data into five-folds. Each classifier gives a classification based on the embeddings of the discourse-text and the discourse-type. The model persuasiveness score, and the model-feedback which is obtained using attention-visualization, combined with the teacher's feedback, eventually goes to the students to help improve efficacy of their arguments.}
\end{figure}

\section{Method}
\subsection{Data Preprocessing and Problem Formulation}

Recall that our goal is to predict the effectiveness rating for each discourse element given its type label. In the training data, except a table of discourse elements, we also have access to the complete essays which these discourse elements are extracted from. Each essay contains a variable number of discourse elements, with possibly repeated type labels.

\subsubsection{Data Preprocessing:} 
In order to fully utilize the context information, when evaluating each discourse element, we aim to include all other discourse elements extracted from the same essay in the input as well. For the purpose of efficiency, predictions are done on the essay level.
\par
The data preprocessing goes as follows. For each essay, we first look for every discourse element extracted from it and locate them within the essay. Then we add special tokens to the beginning and the end of each discourse element indicating the corresponding discourse type. Finally, we concatenate all the discourse elements together to form a new essay, following the same order as when they appear in the original one. An example of preprocessed essays is the following:
\newline
\texttt{
[LEAD\_START] Sometimes having to make a decision is pretty hard thats why we ask for advice from some people. [LEAD\_END] [POSITION\_START] Seeking multiple opinions can help with making a better choice [POSITION\_END] because [CLAIM\_START] \ldots
}
\par
After this step, we tokenize the essays and use the resulting sequences as the inputs to our model.

\subsubsection{Problem Formulation:}
Eventually, we want to produce one prediction, which is a probability distribution over the three different ratings, for every discourse element. Essentially this is a sequence classification problem. However, instead of directly handling it as a sequence classification task, we find that it is more efficacious to formulate the problem as a token classification task. At training time, we label each token with the effectiveness rating of its corresponding discourse element and try to correctly predict all labels. Then at inference time, we take the average scores of overall tokens to obtain a prediction for each discourse element.

\subsection{Model Selection} \label{tab:sec32}

We build our classifier on the pre-trained large language model DeBERTaV3. The DeBERTa model, originally proposed in \cite{he2020deberta}, improves BERT \cite{bert} using a disentangled attention mechanism and an enhanced mask decoder. Then DeBERTaV3 \cite{he2021debertav3} further improves the original DeBERTa model with a new ELECTRA-style pre-training method. We briefly introduce the three techniques here.

\subsubsection{Disentangled Attention}
In a classical attention mechanism, each token is represented by one vector which is the sum of the content embedding and the position embedding, whereas in DeBERTa these two embeddings are kept separate. For each pair of tokens $x_i$ and $x_j$ at position $i$ and $j$ respectively, we have two pairs of embeddings: the content embeddings $\bm{H_i}, \bm{H_j}$ and the relative position embeddings $\bm{P_{i|j}}, \bm{P_{j|i}}$. The cross-attention score between $x_i$ and $x_j$ is then calculated as

\begin{equation}
    A_{i,j} = \bm{H_i} \bm{H_j}^\top + \bm{H_i} \bm{P_{j|i}}^\top + \bm{P_{i|j}} \bm{H_j}^\top,
\end{equation}
where position-to-position attention is omitted in the implementation for lack of useful information.

\subsubsection{Enhanced Mask Decoder:}
DeBERTa is pre-trained using masked language modeling (MLM). The absolute position information is important in this task as well as many other NLP tasks. To use this information, DeBERTa incorporates the absolute position embeddings after the Transformer layers but before the softmax layer for masked token prediction, hence enhancing the mask decoder.

\subsubsection{ELECTRA-style Pre-training:}
DeBERTaV3 replaces the MLM pre-training procedure in DeBERTa with an ELECTRA-style pre-training procedure. In the pre-training stage, DeBERTaV3 trains a generator that aims to minimize an MLM loss and a discriminator which aims to minimize an RTD (Replaced Token Detection) loss simultaneously. A Gradient-Disentangled Embedding Sharing method is adopted to avoid the tug-of-war between the generator and the discriminator.

\subsection{Memory constraints} \label{tab:sec33}
The "scaling laws" draw us towards picking models that are ever larger. Yet, this added performance brought by a larger number of weights does not come for free for all Machine Learning practitioners. 
DeBERTaV3$_{large}$ represents around 800MB of weights when stored in a PyTorch bin file. The virtual machines that a lot of practitioners have access to for free (be it Google collab or Kaggle notebooks) tend to have around 13GB of RAM available. It is not sufficient memory; we lay out a simple memory requirement estimation example below.

For a 1 billion fp32 parameter model we can break down the memory needs as such: 4GB of data just for the weights since these are fp32 numbers, the same amount is needed to store the gradients. On top of this 8 GB, in the case of Adam as a choice of the optimizer (which is the optimizer that we are using) we need to add 8GB for storing the first and second moments of each gradient. Therefore as a rough estimation, we end up with 16GB required just to properly load a 1bn parameter model for training. before taking into account the required memory to store the activations during the forward pass. If we want to load a decently sized model such as DeBERTaV3$_{large}$ we need to make use of a few engineering tricks to circumvent these constraints.

\subsubsection{fp16:} Mixed precision training is the first trick that we used \cite{micikevicius2017mixed}. It is a very intuitive technique that we used which relies on cleverly using low-precision arithmetic. Instead of storing all real numbers in their 32-bit representation, we shift all of them to be represented as 16-bit numbers. It enables saving a lot of memory while not compromising the accuracy of computations (fp16 loses in representative power due to its limited numerical range but has decent precision otherwise). Therefore based on the architecture which uses batch normalization layers we can assume that quantization errors will most likely be negligible since activations are frequently normalized. We can shift the quantization of our neural network by passing "fp16" as an argument to the Trainer object. The implementation that Huggingface uses for quantizing a model does not involve representing all weights, gradients, activations, and moments as fp16 numbers; simply the forward activations saved for gradient computation. Thus it does not halve the memory needs; more optimization techniques are required.

\subsubsection{Gradient checkpointing:} 
Chen et al. introduced this technique - also known as "activation checkpointing" - in their paper \cite{chen2016training}. It uses significantly less memory. When enabled a lot of memory can be freed at the cost of a small decrease in training speed. The memory savings tend to be in the order of $\mathcal{O}(n)$ with n the number of feed-forward layers. The general idea of this technique is to cleverly analyze the computation graph and based on it decide on what results to store. For example, if a low-cost operation of a forward pass can be dropped and only recomputed later during the backward pass it becomes a savvy trade-off between computation and memory.

\subsubsection{Gradient accumulation:} This technique modifies the last step of the backward pass when training a neural network. Instead of updating the gradients after the forward pass and backward pass of each mini-batch, the gradients are saved and the update is only done after several mini-batches. It enables an algorithm to emulate a network training procedure on larger batches even though the execution of the forward and backward pass is done on smaller batches by performing the weights update on their accumulation; hence saving extra memory.

\subsection{Ensembling} 
Ensembling is an approach that combines predictions from multiple models in order to obtain a better predictive performance. There are multiple ways of ensembling models. In our model, we used a "K-fold cross-training" approach - a combination of K-fold cross-validation and bagging approaches, where we divided the training data into five folds and trained five models independently with each fold as a test set. The final prediction is an average of all five models. A brief description of bagging and other ensembling methods is listed below.

\subsubsection{Bagging} We noticed that models trained over different folds of training samples had high variance. To reduce this variance, we used bagging \cite{breiman1996bagging} by averaging the predictions of five different models.

\subsubsection{Boosting} This other paradigm had us use a LightGBM \cite{ke2017lightgbm} Bag-of-Words model \cite{zhang2010understanding}, which is a Bag-of-words classifier with Light Gradient Boosting Method \cite{schapire1999brief}. This model worked decently well, slightly better than a BERT-base sequence classifier, but we needed a stronger model to get a competitive model.

\subsubsection{Stacking} Another ensembling technique that we tried was stacking \cite{pavlyshenko2018using}; we tried to ensemble each of the five models' predictions and the Bag-of-words classifier using another meta-model, which was a neural network. This approach is useful in reducing bias among models. We weren't able to improve the final results using stacking because the five models combined make a strong learner, while the Bag-of-words model is quite weak, therefore the neural network almost completely ignores the Bag-of-words model. It would be interesting to stack a few other weak models before ensembling with our main strong model. We can actually show some improvement in the loss, something that was done by this paper to identify fake news \cite{jiang2021novel}.

\subsection{Hyperparameter Optimization}
As mentioned before DeBERTaV3$_{large}$ is a heavy model. The training time of four models that we then ensemble takes about 12 hours on a Tesla P100. We can understand the limitations that one can have when using free VM access like Colab free or Kaggle notebooks. Hyperparameter optimization was therefore a low priority because of our inability to run instances for more than 30 hours per week. Yet, we decided to focus on one hyperparameter:
the accumulation step interval. We decreased it so that gradients would be updated after every batch at the expense of extra memory usage. It granted us a +0.01 in AUC. 
\subsection{Regularization}
We used several regularization techniques to help our model generalize. We used dropout probabilities of 10\%, 20\%, 30\%, 40\% and 50\% and averaged the outputs to get the final output logits. Additionally, we used Adversarial Weight Perturbation, which is briefly described below. We also tried to implement Scale Invariant Fine-tuning, a work in progress that we also describe below.
\subsubsection{Adversarial Weight Perturbation:} Adversarial Weight Perturbation  \cite{wu2020adversarial},\cite{yu2022robust} is a regularization method that perturbs the weights of the Deep Neural Networks to prevent overfitting to the data. Here, we apply weight perturbation every time the training loss goes below a set threshold and we add noise in the direction of the gradient of the loss function with respect to the weights. This method works similarly to Stochastic Weight Averaging which keeps on trying to push the learner away from falling into local minima. 

\subsubsection{Scale Invariant Fine-tuning:}  Virtual Adversarial Perturbation \cite{miyato2018virtual} is a technique of regularization that introduces small perturbations in the input thus regularizing the model by generating the same output for an example as it would generate for the adversarial perturbation of the example. 
\par
In the case of NLP tasks, these perturbations are added to the word embeddings instead of the original sequence. The problem is that the word embedding values vary largely between different words and models. To solve this, the authors of the DeBERTa paper \cite{he2020deberta} suggest Scale Invariant fine-tuning or SiFT that normalizes embedding layers before applying perturbations. This method significantly improves the performance of the model in the downstream NLP tasks.

\section{Results}

\subsection{Evaluation Metric}
The output is evaluated based on the log loss as follows.
\begin{equation}
    log-loss =  \frac{1}{N} \sum_{i=1}^N \sum_{j=1}^M y_{ij} log(p_{ij})\\    
\end{equation}
where N is the number of rows in the test set, M is the number of class labels, $y_{ij}$ is 1 if observation $i$ is in class $j$ and 0 otherwise, and $p_{ij}$is the predicted probability that observation $i$ belongs to $j$. 

\subsection{Overall Results}
Table \ref{tab:result} below describes the performance of the different language models, that were used for generating the embeddings, for the given task.

\begin{table}[ht]
    \caption{Comparison of performance of various architectures on the persuasiveness prediction task}
    \label{tab:result}
    \centering
\begin{tabular}{lcccr}
\toprule
Language Model & Technique & log-loss  \\
\midrule
$BERT_{base}$    & Fine-tuning & 0.7436  \\
$DeBERTa_{Base}$ & x & 0.7440 \\
$DeBERTa_{large}$ & Fine-tuning and Ensembling & \textbf{0.5806} \\
$DeBERTa_{large}$ & Pseudo-labeling& 0.63626\\ 
$DeBERTa_{large}$ & Fine-tuning w/ tokens& 0.63254 \\
\bottomrule
\end{tabular}
\end{table}

The models trained on DeBERTa Large embeddings significantly outperformed the other models. Section \ref{tab:sec32} explains how DeBERTa improves over the BERT model. Using a larger model improves the performance significantly, however, it also severely increases the computational costs and memory requirements. Section \ref{tab:sec32} explains how we overcame such memory constraints.

\section{Discussion}
This project seeks to identify a way to include Artificial intelligence in assessing argument-persuasiveness. This model, combined with an argument-mining AI, is capable of identifying the sections of an essay that are "effective" or "ineffective" in persuading the reader. Based on the segments identified as "ineffective" by the machine, a teacher can go through those sections carefully, identify the scope of improvements and provide the necessary feedback. It allows to specifically target the segment of the essay that needs attention, thus making the job of the teacher much more specific. It ends up really helping the student identify the sections where they need to strengthen their arguments.

In this project, various aspects of language models were explored in order to achieve competitive accuracy. This discussion section tries to summarize some of the more vital aspects of the architecture, and the key takeaways from those.
\subsection{Choosing the right language model}
The choice of model was a vital piece of the puzzle. A vanilla DeBERTa$_{base}$\cite{he2020deberta} model performs on par with a fine-tuned BERT$_{base}$\cite{bert} architecture which is one of the most popular Transformer-based architectures. Recall that the DeBERTa architecture differs from BERT\cite{bert} or RoBERTa\cite{RoBERT} mainly due to the introduction of the disentangled attention property. Increasing the size of the model by using the DeBERTa$_{large}$ architecture improved the performance significantly. This led us to two major conclusions. Firstly, that the disentangled attention, that separates the token and positional embeddings, creates a more robust representation of text for assessing the its persuasiveness. The AI index report of 2021 \cite{zhang2021ai} shows that the DeBERTa architecture tops the leaderboard of the SuperGLUE benchmark \cite{wang2019superglue} which is a benchmark for complex language understanding tasks. This shows that DeBERTa model, with its disentangled attention mechanism, better encapsulates the contextual understanding of the text and hence, supports the sentence evaluation better than the other Language Models. 

Secondly, a little more trivial conclusion was that incrementing the size of the model actually significantly improves the performance of the model.

\subsection{Ensembling methods}
Among the various methods applied to our initial architecture, ensembling methods applied to DeBERTaV3 led to the most significant improvement. We ensembled five identical DeBERTaV3 architectures each using different training/validation splits of our working dataset. After ensembling these models we reached 0.63 in log loss. However, ensembling requires substantial extra GPU memory during training due to having to deal with four other models. Section \ref{tab:sec33} describes how we went about solving the computational challenges.

We also applied boosting on the bag-of-words model\cite{zhang2010understanding}. Even though a BERT\cite{bert}$_{base}$ model is known to provide significantly better performance for most NLP tasks as compared to a bag-of-words model, results show that boosting improves the bag-of-words model's performance for specific tasks. However, the worsening of performance by stacking shows that ensembling unbalanced models can lead to poorer models. 

\subsection{Overcoming Computational Challenges}
Training a large-language model requires an appropriate quantization of the model. Reducing the precision of the model to fp16 reduced the memory requirements significantly. Furthermore, increasing the gradient accumulation steps reduced the amount of computation required and hence, the amount of time required to train the model. Overall, training a large language model might require making trade-offs between accuracy and training speed, as well as making judgement calls regarding the precision that the model requires.

\section{Future Work}
In future work, we aspire to use Explainable AI (XAI) to have the machine pivot from a grader position to a Teaching Assistant position. The hope is to transform our predictive model into a feedback provider. The feedback would then trigger a conversation between the student, the teacher, and the machine.

XAI is a subset of artificial intelligence that focuses on making the decision-making process of a model transparent and interpretable to human users. In the context of providing feedback to students on the strength of their arguments in an essay, a large language model can be used to analyze the text and then identify key elements such as the structure of the argument, the use of evidence, and the clarity of the writing so that the student can improve on those.

In future work, we aspire to use XAI to make the model's feedback more explainable and use natural language generation (NLG) techniques to generate human-readable explanations as feedback. For example, the model could identify that a student's essay lacks a clear thesis statement and generate the feedback "Your essay does not have a clear thesis statement. A strong thesis statement is essential for guiding the structure and direction of your argument."

XAI would be done by using techniques such as attention visualization, which can show which parts of the text the model is focusing on when making its grade predictions. Once identified, it is shown to the student and thus can help them understand why the machine is giving them a certain grade and how they can improve their writing.

On a more granular level, a way to use attention visualization for feedback is to display a heatmap of the essay, where each word is colored based on the level of attention the model is giving to it. The words that are colored more brightly are the ones that the model is paying more attention to, and therefore are the ones that are most important for the student to focus on when revising their essay. For example, if the model is giving low attention to the introduction, it could be an indication of weak thesis statement or lack of a clear direction for the essay or if the model is giving low attention to certain key vocabulary words related to the topic, it could be an indication of lack of understanding of the topic or weak research.

Overall, attention visualization can be a useful tool for providing feedback to students on their writing by allowing them to see which parts of the essay the model is focusing on and why. This can help them to better understand the model's feedback and make more informed revisions to their writing.

\begin{acknowledgments}
  Thanks to Professor Kilian Weinberger for his support and ideas throughout our work.  
\end{acknowledgments}

\bibliography{sample-ceur}


\end{document}